\documentclass[a4paper,fleqn]{cas-dc}

\usepackage[numbers,sort&compress]{natbib}
\usepackage{amsmath,amssymb}
\usepackage{booktabs,multirow,makecell}
\usepackage[ruled,vlined,linesnumbered]{algorithm2e}
\usepackage[protrusion=true,expansion=false]{microtype}
\usepackage{xurl}

\newcommand{\figfile}[2]{\includegraphics[width=#2]{#1}}

\newcommand{\bw}{\mathbf{w}}

\begin{document}
\let\WriteBookmarks\relax
\def\floatpagepagefraction{0.9}
\def\textpagefraction{.1}

\shorttitle{RBA: a read-out basis adapter for SEEM}

\shortauthors{Al-Ghabban and Mohammadi}

\title [mode = title]{RBA: A Read-out Basis Adapter for Parameter-Efficient Interactive OCTA Vessel Segmentation with SEEM}

\author[1]{Jaafar M. Al-Ghabban}[orcid=0000-0003-3562-0034]

\ead{jafar.alghban@gmail.com}

\credit{Conceptualization, Methodology, Software, Validation, Formal analysis, Investigation, Data curation, Visualization, Writing -- original draft}

\author[1]{Mohammad Reza Mohammadi}[orcid=0000-0002-1016-9243]

\cormark[1]

\ead{mrmohammadi@iust.ac.ir}

\credit{Conceptualization, Methodology, Supervision, Project administration, Writing -- review \& editing}

\affiliation[1]{organization={School of Computer Engineering, Iran University of Science and Technology},
            city={Tehran},
            country={Iran}}

\cortext[1]{Corresponding author}

\begin{abstract}
Retinal disease monitoring with optical coherence tomography angiography (OCTA) relies on segmentation of vessels and the foveal avascular zone (FAZ). Promptable models such as SEEM (Segment Everything Everywhere All at Once) allow click-based refinement, but without adaptation SEEM segments thin vessels poorly and needs many clicks. Fully fine-tuning its 342 million parameters on a few hundred images is costly and risks overfitting, and parameter-efficient methods such as low-rank adaptation (LoRA) usually adapt only the encoder. Yet SEEM reads every candidate mask from one feature map at quarter resolution, too coarse for vessels one to three pixels wide. This study introduces the Read-out Basis Adapter (RBA), whose read-out branch generates 32 image-specific basis maps at half resolution. Each candidate accesses them through its own vector and a zero-initialised projection, so the model starts exactly as SEEM. The encoder branch, a standard rank-1 LoRA, supplies the information needed to tell arteries from veins. RBA adds 0.47 million trainable parameters. Under simulated error-driven clicks, with the ground truth used only to place them, RBA outperforms LoRA, Conv-LoRA and AdaptFormer on OCTA-500 3~mm large vessels. Averaged over three seeds, it reaches an intersection over union (IoU) of 85.16\% and a Dice score of 91.93\%, 2.90 and 1.72 percentage points above Conv-LoRA. On the 6~mm subset, one run gives margins of 4.35 and 2.76 points. The gain extends to arteries, veins and ROSE-O but not to the FAZ. In SEEM, adaptation capacity pays more at the mask read-out than in the encoder.
\end{abstract}



\maketitle

\section{Introduction}\label{sec:intro}

Retinal vessel segmentation underpins the quantitative study of retinal diseases such as diabetic retinopathy and glaucoma. Diabetic retinopathy affects more than one in five people with diabetes, and 160.5 million adults are projected to have it by 2045~\cite{teo2021global}. Glaucoma affected an estimated 64.3 million people aged 40 to 80 in 2013, a number projected to reach 111.8 million by 2040~\cite{tham2014global}. Both diseases change the retinal microvasculature, so vessel density, vessel calibre and the shape of the foveal avascular zone (FAZ) are used to study them. Each of these measurements is only as reliable as the segmentation behind it.

Optical coherence tomography angiography (OCTA) images the retinal microvasculature in depth-resolved scans. Unlike fluorescein angiography, it needs no dye injection~\cite{de2015review}. The measurements above are taken on its en-face projections, which show the vessels of selected retinal layers.

Deep networks trained on dense labels now dominate this segmentation. Public datasets such as OCTA-500~\cite{li2024octa} and ROSE~\cite{ma2020rose} have turned it into a benchmark. On these datasets, Li et al. with IPN~\cite{li2020image}, Hu et al. with Joint-Seg~\cite{hu2022joint} and Peng et al. with FARGO~\cite{peng2021fargo} report high scores. These networks share two limits. Each new target or imaging device needs new dense labels, and an error can only be corrected by editing the mask pixel by pixel.

Three properties of OCTA projections make the segmentation hard. Many vessels on the 3~mm OCTA-500 images are one to three pixels wide, and vessels below three pixels make up about a quarter of the large-vessel label. Huang et al.~\cite{huang2026gaze} report that vessel ends are faint and low in contrast, and Hu et al.~\cite{hu2025dual} address the noise and artefacts that the projections carry. Vessel pixels are also a small minority of each image, so quality is decided on thin branches and vessel edges, whereas the FAZ is a single compact region.

Interactive segmentation addresses both limits of dedicated networks. The user clicks on a missed or a false region, and the model revises its mask after each click, as in FocalClick, SimpleClick and RITM~\cite{chen2022focalclick,liu2023simpleclick,sofiiuk2022reviving}. Promptable foundation models such as the Segment Anything Model (SAM)~\cite{kirillov2023segment} and SEEM, short for Segment Everything Everywhere All at Once~\cite{zou2023segment}, offer this mode for general images. SAM-OCTA~\cite{wang2024sam,chen2025samocta} and SAM-OCTA2~\cite{chen2025sam} adapted SAM to OCTA, but with prompts drawn from the ground truth. Out of the box, these models do not see OCTA vessels well. With a frozen encoder and a trained decoder, SEEM reaches a Dice score of only 75.83\% on large vessels and 50.98\% on veins in this study, although its pixel accuracy is 96.6\%. Even with its decoder trained, this model did not reach 90\% Dice within the 20 allowed clicks on the large-vessel test images, so SEEM must be adapted to OCTA. Fine-tuning all of its 341.7~M parameters on a few hundred labelled images is costly and risks overfitting, which motivates parameter-efficient adaptation.

\begin{figure*}[!t]
\centering
\figfile{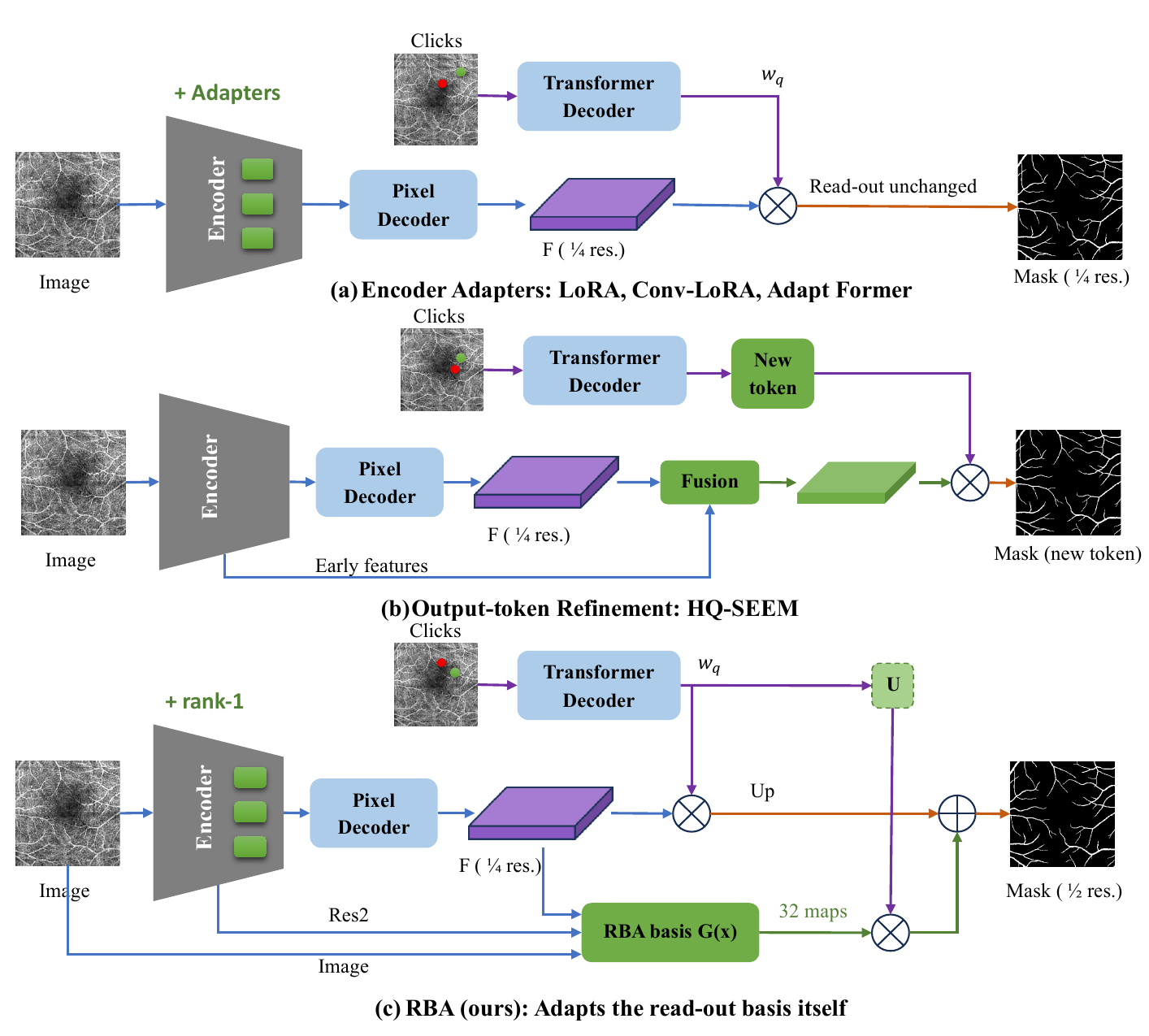}{0.72\textwidth}
\caption{Three ways to adapt a promptable segmentation model: (a) encoder adapters such as LoRA, Conv-LoRA and AdaptFormer, (b) output-token refinement as in HQ-SEEM, and (c) RBA, which adapts the read-out basis of SEEM. Green marks trainable additions.}
\label{fig:families}
\end{figure*}

The first group of adaptation methods changes how the image is encoded, as panel a of Fig.~\ref{fig:families} shows. Hu et al.~\cite{hu2021lora} proposed low-rank adaptation (LoRA), which adds a trainable low-rank update to frozen weights, and Zhong et al.~\cite{zhong2024convolution} placed a convolution inside this low-rank bottleneck in Conv-LoRA. Such encoder adapters change what the feature map of SEEM contains. They change neither its quarter resolution nor the way masks are read from it.

The second group changes how the final mask is produced, as panel b of Fig.~\ref{fig:families} shows. Ke et al.~\cite{ke2023segment} proposed HQ-SAM, which adds a high-quality output token whose mask, computed from fused early and late encoder features, replaces the mask of SAM. This token sharpens the mask, but it bypasses the candidates and the selection among them on which interaction relies.

Neither group acts at the step at which SEEM forms its masks. Each decoder query of SEEM produces a vector, and its candidate mask is the dot product of this vector with one feature map at a quarter of the input resolution. Every mask that SEEM can produce is therefore a combination of this coarse basis. No existing adapter acts at this read-out while keeping the candidates and the selection of SEEM and starting from the exact pretrained model.

To address this gap, this study proposes the Read-out Basis Adapter (RBA), shown in panel c of Fig.~\ref{fig:families}. RBA consists of two branches. The read-out branch builds 32 basis maps at half the input resolution from the image alone. Every candidate reads them through its own vector via a linear map that starts at zero, like the update of LoRA~\cite{hu2021lora}. The encoder branch is a standard rank-1 LoRA that supplies the information needed to tell arteries from veins. RBA therefore starts exactly as SEEM, keeps its candidates and selection, and computes the new basis once per image, so a click costs almost nothing more. Fig.~\ref{fig:overview} shows the complete pipeline.

The contributions of this work are fourfold. First, it proposes RBA, an adapter of the basis from which every SEEM candidate mask is read, with 0.47~M parameters, which is 0.14\% of SEEM. Second, it shows that in SEEM adaptation capacity pays more at the read-out than in the encoder. Once the read-out is adapted, a rank-1 encoder branch suffices, and the read-out improves every encoder adapter to which it is added. Third, it evaluates the model with simulated clicks in which no ground truth enters the prediction, and it tests every difference in pairs under decision rules fixed before the runs. Fourth, it reports results on OCTA-500 and ROSE-O. On large vessels of the 3~mm subset, RBA outperforms LoRA, Conv-LoRA and AdaptFormer on all 50 test images, as Table~\ref{tab:adapters} reports. Averaged over three training seeds, it exceeds Conv-LoRA by 1.72 percentage points in Dice and 2.90 in intersection over union (IoU), and the seed-to-seed standard deviation of Dice is 0.02\% for both models. On the 6~mm subset the Dice margin grows to 2.76 percentage points, and the gain extends to veins, arteries and ROSE-O, whereas the FAZ shows no significant difference.

The rest of this paper is organised as follows. Section~\ref{sec:related} reviews related work, Section~\ref{sec:method} presents the methodology, Section~\ref{sec:results} reports the results and Section~\ref{sec:conclusion} concludes.

\section{Related work}\label{sec:related}

\subsection{Deep learning for OCTA vessel segmentation}

Progress in OCTA vessel segmentation has relied on public datasets with pixel-level labels. Li et al.~\cite{li2024octa} released OCTA-500, which covers 500 eyes at 3~mm and 6~mm fields of view with pixel-level labels for large vessels, arteries, veins, capillaries and the FAZ. Ma et al.~\cite{ma2020rose} introduced ROSE, which adds vessel labels from two devices, together with the OCTA-Net model. Hao et al.~\cite{hao2022retinal} extended ROSE with FAZ labels as ROSE-O and proposed the multitask network VAFF-Net.

The dedicated networks differ mainly in how they build vessel structure into the model. Li et al.~\cite{li2020image} proposed the image projection network IPN, which learns the projection from the 3D volume to the en-face image. Peng et al.~\cite{peng2021fargo} with FARGO and Hu et al.~\cite{hu2022joint} with Joint-Seg segment the FAZ and the vessels as a joint task, and Melo et al.~\cite{melo2026multitask} applied multitask learning to the FAZ. Chen et al.~\cite{chen2024snake} proposed SSW-OCTA, which uses snake-like convolutions that follow vessel paths, and Ma et al.~\cite{ma2026deep} combined full-scale skip connections with split attention in RS\_Unet3+. Two further networks address the properties that make vessels hard to segment. Hu et al.~\cite{hu2025dual} disentangle vessels from noise and artefacts in D2Net, and Huang et al.~\cite{huang2026gaze} add patch-wise attention for faint vessel ends in LSENet.

These networks reach high accuracy, but they share two costs. Every new target or imaging device needs new dense labels, and an error in a prediction can only be corrected by editing the mask pixel by pixel.

Promptable models reached OCTA through SAM-OCTA of Wang et al.~\cite{wang2024sam} and Chen et al.~\cite{chen2025samocta} and through SAM-OCTA2 of Chen et al.~\cite{chen2025sam}, which fine-tune SAM and SAM~2 with LoRA. Their prompts are sampled from the ground truth, in SAM-OCTA from every connected component of the label. This measures how well a model segments when it receives ideal prompts, not how it behaves when a user corrects its mistakes. The present work evaluates the second case, with error-driven clicks and the model's own choice of candidate.

\subsection{Promptable models and parameter-efficient adaptation}

Interactive segmentation follows a standard protocol, used by Sofiiuk et al.~\cite{sofiiuk2022reviving} in RITM, by Chen et al.~\cite{chen2022focalclick} in FocalClick and by Liu et al.~\cite{liu2023simpleclick} in SimpleClick. A simulated user clicks at the centre of the largest error region, and the score is the number of clicks needed to reach a target IoU or Dice. The ground truth places the clicks but never chooses the output.

Kirillov et al.~\cite{kirillov2023segment} extended prompting to many prompt types with SAM, and Zou et al.~\cite{zou2023segment} did so with SEEM, whose decoder derives from the Mask2Former of Cheng et al.~\cite{cheng2022masked}. Evaluations that pick the candidate or the threshold against the ground truth overstate what a model achieves on its own, so this work leaves both choices to the model.

Beyond LoRA~\cite{hu2021lora} and Conv-LoRA~\cite{zhong2024convolution}, introduced in Section~\ref{sec:intro}, several variants change what happens inside the low-rank bottleneck. Chen et al.~\cite{chen2025lora} proposed NAS-LoRA, which searches the bottleneck structure, and Yin et al.~\cite{yin20231} proposed LoRand for dense prediction. Iltaf et al.~\cite{iltaf2025vesselsam} used atrous convolutions for aortic vessels in VesselSAM. Bottleneck adapters insert small multilayer perceptrons (MLPs) instead. Chen et al.~\cite{chen2022adaptformer} placed them in parallel with each MLP in AdaptFormer, and Chen et al.~\cite{chen2023sam} introduced SAM-Adapter. Wu et al.~\cite{wu2025medical} proposed the Medical SAM Adapter, and Hu et al.~\cite{hu2025spa} added spatial priors in SPA. Closest to the vessel setting, Khazem~\cite{khazem2026topolora} combined LoRA with a lightweight spatial convolutional adapter on high-resolution embeddings and a clDice loss in TopoLoRA-SAM, for non-interactive segmentation of thin structures such as retinal vessels. For SEEM itself, Abou Baker et al.~\cite{abou2024parameter} compared adapters and LoRA for instance segmentation, with SEEM and Mask DINO as base models.

These methods assume that segmentation lacks a better representation. For SEEM this is only half of the problem, since better encoder features still reach the masks through the same quarter-resolution read-out. The case for convolutional adapters is weaker still, because the FocalNet encoder of SEEM~\cite{yang2022focal} already aggregates context with depth-wise convolutions. On the output side, HQ-SAM~\cite{ke2023segment} sharpens the mask with an extra token, but it bypasses the candidates and the selection that interaction relies on.

In summary, encoder adapters change what the features contain, and output-token methods replace the candidates, while the basis from which SEEM reads its candidates stays unchanged. RBA combines the zero-initialised low-rank update of LoRA with high-resolution image features and applies them to this read-out. The next section formalises the read-out and presents RBA.

\section{Methodology}\label{sec:method}

\subsection{Problem formulation}\label{sec:problem}

In interactive segmentation, the model refines its mask over several rounds of user clicks. Let $x\in\mathbb{R}^{3\times S\times S}$ be the OCTA input, three stacked en-face projections resized to $S=512$. Let $y\in\{0,1\}^{H\times W}$ be its label on the native grid, where $H=W=304$ for the 3~mm subset and ROSE-O and $H=W=400$ for the 6~mm subset. An interactive session is a sequence of rounds $k=1,\dots,K$. In round $k$ the user has placed the clicks $c_{1:k}$, each with a location and a sign, positive on a missed vessel and negative on a false one, and the model returns a mask $\hat{y}_k$. A good model reaches a high Dice in few rounds while selecting its own output at a fixed threshold. The ground truth is used only to simulate where a user would click.

SEEM processes the input with a FocalNet-L encoder~\cite{yang2022focal} and a pixel decoder, which produce a mask-feature map $F$ of size $C\times S/4\times S/4$ with $C=512$. Given the prompts, the transformer decoder updates $Q=101$ queries, and the mask-embedding MLP of SEEM maps each query to a vector $\bw_q$ of dimension $C$. Following the prototype-based view of instance segmentation~\cite{bolya2019yolact}, the logit of candidate $q$ at pixel $u$ of the $S/4\times S/4$ grid is
\begin{equation}
m_q(u) = \langle \bw_q, F(u)\rangle = \sum_{c=1}^{C} w_{q,c}\,F_c(u).
\label{eq:readout}
\end{equation}
This dot product is the read-out of SEEM. SEEM selects one candidate per click with a query pooled at the click. A click changes the vectors $\bw_q$ and the selected candidate, but not $F$. For a given image, every candidate is therefore the sigmoid of a weighted sum of the same 512 maps at $128\times128$ pixels, for a label grid of $304\times304$. If the shape of a thin vessel or of a vessel edge is not in the span of these maps, no candidate can express it.

\subsection{Read-out Basis Adapter}\label{sec:rba}

\begin{figure*}[!t]
\centering
\figfile{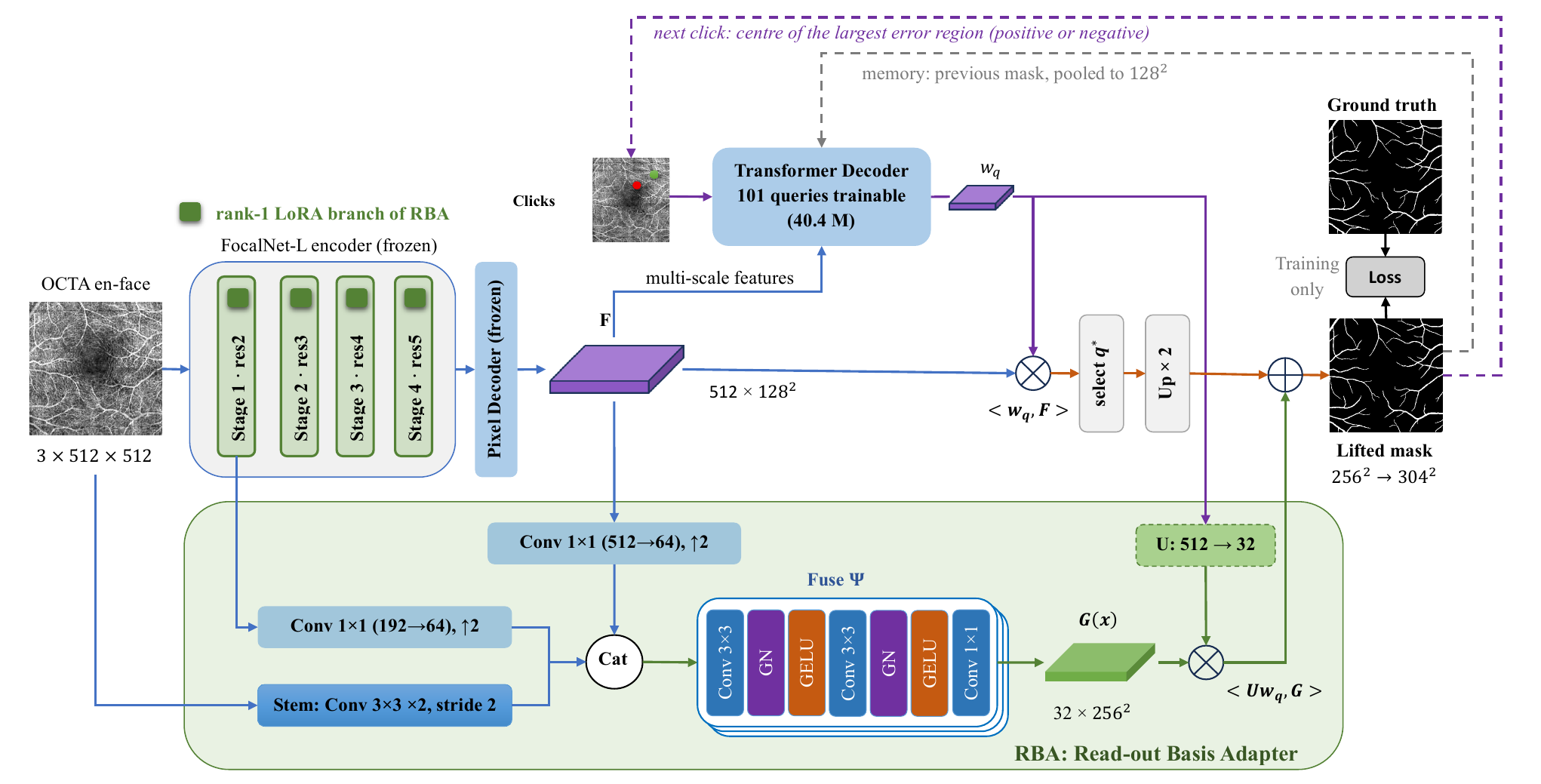}{\textwidth}
\caption{Overview of SEEM with RBA. Blue marks the image path, purple the click path and green the read-out term of RBA. Dashed loops mark the next click and the previous-mask memory.}
\label{fig:overview}
\end{figure*}

RBA is built on SEEM and adds two branches. SEEM with the FocalNet-L encoder has 341.7~M parameters in total. The FocalNet-L encoder, the pixel decoder that produces $F$ and the language encoder, which the click task does not use, stay frozen. The transformer decoder, with 39.8~M parameters, is trained, and its previous-mask memory and its selection rule are used unchanged. A learned positional click prompt with 0.57~M parameters is trained with the decoder. It combines the click tokens of SEEM with a learned positional encoding of each click and is shared by every configuration compared in this work. RBA itself has 474,168 parameters, of which 270,840 belong to the encoder branch and 203,328 to the read-out branch. The image path runs once per image, and only the decoder and the read-out run in every click round. Fig.~\ref{fig:overview} shows how the branches connect to SEEM.

The encoder branch supplies the small, task-specific change that the frozen encoder needs to tell vessel types apart. It is a standard rank-1 LoRA~\cite{hu2021lora} in the linear layers $f$, $\mathit{fc1}$, $\mathit{fc2}$ and $\mathit{proj}$ of all 24 FocalNet-L blocks, which makes 96 layers:
\begin{equation}
y = W h + \frac{\alpha}{r}\, B A h, \qquad r = 1, \quad \frac{\alpha}{r} = 2.
\label{eq:lora}
\end{equation}
Here $W$ is the frozen weight, $h$ is the layer input, $A\in\mathbb{R}^{1\times d_{\mathrm{in}}}$, $B\in\mathbb{R}^{d_{\mathrm{out}}\times 1}$, and $B$ is initialised to zero. The rank is kept at one because a larger encoder adapter does not improve large-vessel segmentation once the read-out is adapted, as Section~\ref{sec:adapters} shows.

\begin{figure*}[!t]
\centering
\figfile{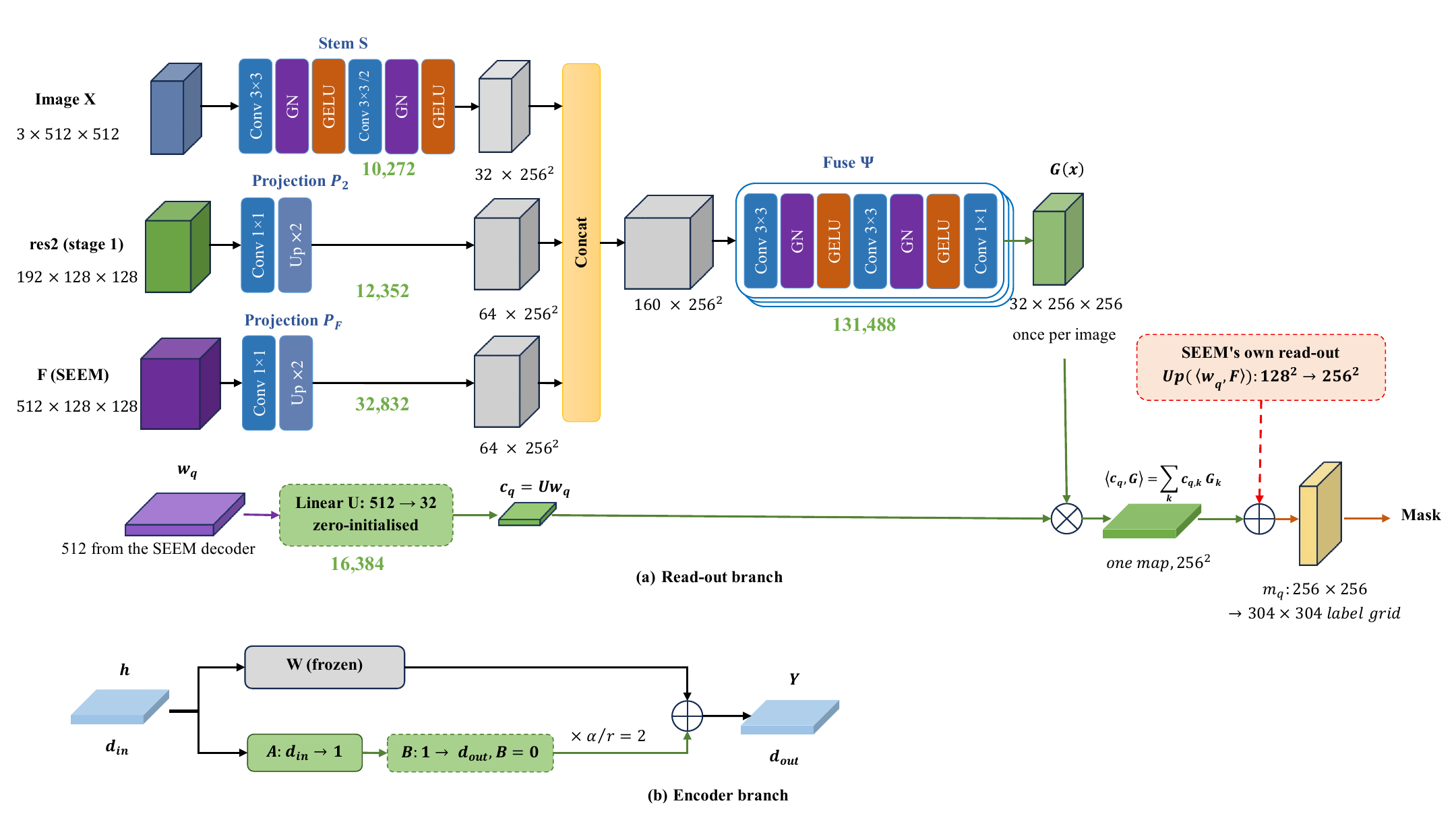}{\textwidth}
\caption{The RBA module: (a) read-out branch and (b) encoder branch. Green numbers give parameter counts, dashed green boxes mark zero initialisation, and red marks the unchanged read-out of SEEM.}
\label{fig:module}
\end{figure*}

The read-out branch supplies the spatial detail that the quarter-resolution features $F$ lack. It builds $n_b=32$ maps at half the input resolution from three sources of increasing depth:
\begin{equation}
G(x) = \Psi\big( [\, \mathcal{S}(x);\ \mathrm{Up}(P_2\,\mathrm{res2});\ \mathrm{Up}(P_F\,F) \,] \big).
\label{eq:basis}
\end{equation}
Here $[\cdot\,;\cdot]$ denotes channel concatenation, $\mathrm{Up}$ denotes bilinear upsampling by two, and $G(x)\in\mathbb{R}^{n_b\times S/2\times S/2}$. The stem $\mathcal{S}$ processes the image rescaled to $[-1,1]$ with two $3\times3$ convolutions, the second with stride 2, each followed by group normalisation and a Gaussian error linear unit (GELU). $P_2$ and $P_F$ are $1\times1$ projections to 64 channels. $P_2$ takes the stage-1 features res2 of FocalNet-L, which carry early edges and textures, and $P_F$ takes $F$, the semantics that SEEM already uses. The fusion network $\Psi$ merges the 160 concatenated channels with two $3\times3$ convolutions, each followed by group normalisation and GELU, and a $1\times1$ convolution to the 32 maps. $G$ depends on the image alone and is computed once per image, before any click. Fig.~\ref{fig:module} gives the layers of both branches and their parameter counts.

The lifted read-out lets every candidate use the new basis without changing the mechanism of SEEM. RBA adds a second, zero-initialised term to Eq.~\eqref{eq:readout}:
\begin{equation}
m_q = \mathrm{Up}\big( \langle \bw_q, F \rangle \big) + \langle U \bw_q,\, G(x) \rangle .
\label{eq:lift}
\end{equation}
Here $U\in\mathbb{R}^{n_b\times C}$, with 16,384 parameters, maps each query vector to $n_b$ coefficients. Equivalently, the read-out basis becomes $\mathrm{Up}(F)+U^{\top}G(x)$, a rank-$n_b$ update of the basis, in the same way as the term $BA$ in LoRA is a low-rank update of a weight matrix.

Four properties follow. First, the start is exact. Since $U$ is zero at the start, $m_q$ equals $\mathrm{Up}(\langle \bw_q, F\rangle)$, the read-out of the pretrained model, because bilinear upsampling is linear and commutes with the sum over channels. Second, directions are kept. For any $\bw$ in the null space of $U$ the mask is the stock mask, so RBA re-assigns at most $n_b$ of the 512 read-out directions. Third, the branch depends on the image only. A click therefore costs one $512\rightarrow32$ linear map and a weighted sum of 32 maps of $256\times256$ pixels, which is about two million multiply-adds. Fourth, learning is ordered. With $U=0$ the gradient of $m_q$ with respect to $G$ is zero, so only $U$ learns in the first step, and the layers that build $G$ follow once $U$ has moved. The first, second and fourth properties were verified numerically at the tensor sizes of SEEM.

\subsection{Click simulation and training}\label{sec:training}

Clicks are simulated from the ground truth, which a real user does not have. The first click is placed at the maximum of the distance transform of the label, inside the thickest vessel. Each later click follows the error-driven rule of Sofiiuk et al.~\cite{sofiiuk2022reviving}. The missed and the false pixels of the current mask at threshold 0.5 form two error maps, and a $7\times7$ window around every earlier click is excluded from both. The click goes to the centre of the thickest error region, that is, to the larger of the two distance-transform maxima. It is positive on a missed region and negative on a false one, and a random error pixel is used if no error pixel remains outside the excluded windows. Clicks are placed on the label grid, mapped to the input and rendered as disks of radius 8 pixels on the $512\times512$ input. Between rounds, the lifted mask is area-pooled to $128\times128$, detached from the gradient and passed to SEEM as the previous-mask memory. This mask belongs to the supervised candidate during training and to the candidate selected by SEEM at test time.

During training, all candidates are lifted and the one that best matches the label is supervised. The mask loss of round $k$ combines Dice~\cite{milletari2016v}, focal~\cite{lin2017focal} and Tversky~\cite{salehi2017tversky} terms on the lifted mask, resized to the native grid:
\begin{equation}
\mathcal{L}^{(k)} = 0.4\,\mathcal{L}_{\mathrm{D}}^{(k)} + 0.3\,\mathcal{L}_{\mathrm{F}}^{(k)} + 0.3\,\mathcal{L}_{\mathrm{T}}^{(k)}.
\label{eq:roundloss}
\end{equation}
The focal parameters are $\alpha=0.25$ and $\gamma=2$, and the Tversky weights are 0.3 for false positives and 0.7 for false negatives, which favours the recall of thin vessels. A second term penalises a round that is worse than the previous one:
\begin{equation}
\mathcal{L}_{\mathrm{impr}} = \frac{1}{K-1} \sum_{k=2}^{K} \Big[ \mathcal{L}_{\mathrm{D}}^{(k)} - \mathrm{sg}\big( \mathcal{L}_{\mathrm{D}}^{(k-1)} \big) \Big]_{+} ,
\label{eq:impr}
\end{equation}
where $[\cdot]_{+}=\max(0,\cdot)$ and $\mathrm{sg}$ stops the gradient. The total loss is
\begin{equation}
\mathcal{L} = \frac{1}{K} \sum_{k=1}^{K} \mathcal{L}^{(k)} + 0.5\, \mathcal{L}_{\mathrm{impr}} .
\label{eq:total}
\end{equation}
Algorithm~\ref{alg:train} summarises the training loop.

\begin{algorithm}[!htb]
\small
\caption{Training SEEM with RBA and simulated clicks.}\label{alg:train}
\KwIn{training set $\mathcal{D}$; SEEM with frozen encoder and pixel decoder; trainable set $\Theta$: SEEM decoder, encoder branch $(A,B)$, read-out branch $(\mathcal{S},P_2,P_F,\Psi,U)$; rounds $K$; epochs $E$}
\KwOut{trained model $f_{\Theta^\star}$}
Initialise $B\leftarrow 0$, $U\leftarrow 0$ \tcp*{the model starts as SEEM}
\For{$\mathit{epoch}=1,\dots,E$}{
  \ForEach{image $x$ with label $y$ in $\mathcal{D}$, with dihedral and intensity augmentation}{
    $(F,\mathrm{res2})\leftarrow$ encoder and pixel decoder$(x)$; $G\leftarrow G(x)$ \tcp*{once per image}
    $c_1\leftarrow$ distance-transform peak of $y$; $\mathit{memory}\leftarrow\emptyset$\;
    \For{$k=1,\dots,K$}{
      $\{\bw_q\}_{q=1}^{Q}\leftarrow$ SEEM decoder$(F, c_{1:k}, \mathit{memory})$\;
      $m_q\leftarrow \mathrm{Up}(\langle\bw_q,F\rangle)+\langle U\bw_q,G\rangle$ for all $q$ \tcp*{Eq.~\eqref{eq:lift}}
      $q^{\dagger}\leftarrow\arg\max_q \mathrm{Dice}(m_q,y)$\tcp*{training only}
      accumulate $\mathcal{L}^{(k)}(m_{q^{\dagger}},y)$ \tcp*{Eq.~\eqref{eq:roundloss}}
      $\mathit{memory}\leftarrow$ area-pool$(m_{q^{\dagger}})$; $c_{k+1}\leftarrow$ error-driven click\;
    }
    $\mathcal{L}\leftarrow \frac{1}{K}\sum_k\mathcal{L}^{(k)}+0.5\,\mathcal{L}_{\mathrm{impr}}$ \tcp*{Eqs.~\eqref{eq:impr}, \eqref{eq:total}}
    $\Theta\leftarrow\Theta-\mathrm{AdamW}(\nabla_{\Theta}\mathcal{L})$\;
  }
}
\Return{$f_{\Theta^\star}$, the checkpoint with the best validation Dice@4}
\end{algorithm}

\subsection{Datasets and implementation}\label{sec:data}

The experiments use two public datasets, OCTA-500 with its 3~mm and 6~mm subsets and ROSE-O. The OCTA-500 3~mm subset~\cite{li2024octa} contains 200 eyes with en-face projections of $304\times304$ pixels, split by the dataset authors into 140 training, 10 validation and 50 test images, which are images 10301--10440, 10441--10450 and 10451--10500. The input stacks three projections, namely the full-depth projection, the projection from the internal limiting membrane to the outer plexiform layer (ILM--OPL) and the projection from the outer plexiform layer to Bruch's membrane (OPL--BM). The stack is resized to $512\times512$ and normalised with the pixel statistics of SEEM. The 6~mm subset contains 300 eyes with projections of $400\times400$ pixels over a $6\times6$~mm field of view, so that a vessel spans about 1.5 times fewer pixels. It uses the split of the dataset paper, with 240 training, 10 validation and 50 test images, which are images 10001--10240, 10241--10250 and 10251--10300, and the same three projections.

ROSE-O~\cite{hao2022retinal} extends ROSE~\cite{ma2020rose} with FAZ labels. It provides en-face projections of the superficial, deep and inner vascular complexes (SVC, DVC and IVC) of 39 subjects at $304\times304$ pixels. The vessel and FAZ labels are drawn on the IVC projection. It was acquired at a different centre and in a different population from OCTA-500, namely patients with Alzheimer's disease and controls, but with a device of the same family, the Optovue RTVue XR Avanti. The released split gives 30 training and 9 test subjects, and five training subjects, drawn with a fixed seed, are held out for validation.

On the 3~mm subset, four targets are studied. Large vessels are the main target, veins and arteries split the large-vessel labels into the two vessel types, and the FAZ, a single compact region, serves as a non-tubular control. On the 6~mm subset and on ROSE-O, RBA is compared with Conv-LoRA, the strongest encoder adapter on the 3~mm subset. Capillary labels are used only in the comparison with an alternative read-out design in Section~\ref{sec:ablation}.

All models are implemented in PyTorch on the public FocalNet-L checkpoint of SEEM. The encoder and the pixel decoder are frozen and run in evaluation mode. The decoder of SEEM, with 40.4~M parameters including the positional prompt, and the adapter are trained, except in the frozen-SEEM experiment of Section~\ref{sec:adapters}. The optimiser is AdamW~\cite{loshchilov2017decoupled} with a weight decay of $10^{-4}$. The learning rates are $5\times10^{-5}$ for the decoder, $2\times10^{-4}$ for encoder adapters and $5\times10^{-4}$ for new modules, namely the click prompt and the read-out branch. They follow a three-epoch linear warm-up and a cosine decay to 1\% of their value. Gradients are clipped at norm 1, the batch size is 1, and training runs for 100 epochs with $K=4$ click rounds per image. An exponential moving average of the weights with decay 0.999 is used for validation, for the saved checkpoint and for the test. Training images are augmented with the eight dihedral transforms and, with probability 0.8, a gamma change of intensity with a small change of contrast and brightness. The checkpoint with the best validation Dice@4 is kept, at the threshold that is best on the validation set, and every test result uses the fixed threshold 0.5. To measure training variation, the main comparison on 3~mm large vessels, RBA against LoRA with rank 16 and Conv-LoRA, was repeated with three training seeds, 42, 1 and 2. These rows are reported as mean $\pm$ standard deviation over the seeds, and all other rows come from seed 42. All experiments ran on a laptop with an Intel Core i7 processor, 32~GB of RAM and a single NVIDIA GeForce RTX 3070 Ti Laptop GPU with 8~GB, with PyTorch 2.1 and CUDA 12.1. One epoch of RBA takes about 4.8~min, against about 6.4~min for Conv-LoRA.

The compared adapters share the pipeline, the 100 epochs and the training of the decoder of SEEM, and differ only in the adapter. LoRA is used with rank 1 and rank 16, and Conv-LoRA with rank 16 in its single-scale form, with one depth-wise $3\times3$ convolution inside the bottleneck. AdaptFormer uses a bottleneck of 12 channels in all 24 blocks, which matches the parameter budget of RBA, and its output scale is set to $\alpha/r=2$ as for the LoRA rows. The RBA read-out is also added to LoRA with rank 16 and to Conv-LoRA, and the read-out branch is trained alone. All baselines are re-implementations in this pipeline.

\subsection{Evaluation protocol and metrics}\label{sec:eval}

All results use a protocol in which no ground truth enters the prediction. The model selects its own candidate with the rule of SEEM, the threshold is fixed at 0.5, and the ground truth only places the simulated clicks. Predictions are resized bilinearly to the native label grid, which is $304\times304$ for the 3~mm subset and ROSE-O and $400\times400$ for the 6~mm subset, and are scored there.

For a binary prediction $P$ and a label $Y$, the Dice score is
\begin{equation}
\mathrm{Dice}(P, Y) = \frac{2\,|P \cap Y|}{|P| + |Y|}.
\label{eq:dice}
\end{equation}
The primary measure, fixed before the runs, is the Dice score after four clicks, Dice@4. The centre-line measure clDice~\cite{shit2021cldice} compares the skeletons $S_P$ and $S_Y$ of the prediction and the label:
\begin{equation}
\begin{aligned}
\mathrm{clDice} &= \frac{2\,T_{\mathrm{prec}}\,T_{\mathrm{sens}}}{T_{\mathrm{prec}} + T_{\mathrm{sens}}},\\
T_{\mathrm{prec}} &= \frac{|S_P \cap Y|}{|S_P|}, \qquad T_{\mathrm{sens}} = \frac{|S_Y \cap P|}{|S_Y|}.
\end{aligned}
\label{eq:cldice}
\end{equation}
It rewards connected vessel trees. Click effort is the mean number of clicks (NoC) needed to reach a target Dice $\tau$ over the $N$ test images:
\begin{equation}
\mathrm{NoC}@\tau = \frac{1}{N} \sum_{i=1}^{N} \min\{\, k \le 20 : \mathrm{Dice}_k^{(i)} \ge \tau \,\}.
\label{eq:noc}
\end{equation}
An image that never reaches $\tau$ counts as 20 clicks, and NoC@90 is reported. Following the OCTA-500 paper~\cite{li2024octa}, intersection over union, accuracy, sensitivity and specificity are also reported, abbreviated IoU, ACC, SE and SP. All scores are given in percent, and differences between scores are absolute differences in percentage points, written with \%.

Rows are compared image by image on the same test images, which are 50 images for OCTA-500 and 9 subjects for ROSE-O. Each comparison reports the mean difference with a 95\% $t$-interval and the number of images that improved. It also reports a two-sided Wilcoxon signed-rank test~\cite{wilcoxon1945individual}, corrected by the Holm method~\cite{holm1979simple} within each family of comparisons and target. For the rows with three seeds, these paired tests use seed 42, and the per-seed differences are reported as well. The decision rules were fixed before the runs. A row B beats a row A if the difference is positive and the corrected $p$-value is below 0.05. B matches A if the difference is not significant and the whole interval lies within $\pm0.3\%$. Otherwise the result is unclear. Rows are compared only if they were trained for the same number of epochs.

\section{Results}\label{sec:results}

\subsection{Comparison with encoder adapters}\label{sec:adapters}

The first experiment compared adaptation at the read-out with adaptation at the encoder on large vessels of the OCTA-500 3~mm subset. Nine configurations were trained under identical conditions. They were SEEM without an adapter, LoRA with rank 1 and rank 16, Conv-LoRA, AdaptFormer, the read-out branch alone, RBA, and Conv-LoRA and LoRA with rank 16 combined with the RBA read-out. Table~\ref{tab:adapters} lists the parameter count, Dice@4, clDice, sensitivity, specificity and NoC@90 of each configuration, and the paired difference between RBA and each row.

\begin{table*}[!t]
\centering
\caption{Adapters on large vessels, OCTA-500 3~mm subset, 50 test images, scores in \%. Enc. and Read-out mark the two branches. Rows marked $^{\ddagger}$ give the mean $\pm$ standard deviation over three training seeds; the other rows use seed 42. The last two columns give the paired change from the row to RBA for seed 42, with its 95\% interval, and the number of images on which RBA is better; $p<10^{-10}$ except $^{\dagger}$ ($p=0.012$). Bold: best; underlined: second best.}
\label{tab:adapters}
\footnotesize
\setlength{\tabcolsep}{3.4pt}
\resizebox{\textwidth}{!}{%
\begin{tabular}{lcccccccccc}
\toprule
Method & Enc. & Read-out & Params & Dice@4 & clDice & SE & SP & NoC@90 & RBA $-$ row [95\% CI] & RBA better \\
\midrule
No adapter & -- & -- & 0 & 75.83 & 80.44 & 80.05 & 97.79 & 20.00 & -- & -- \\
LoRA $r{=}1$ & \checkmark & -- & 0.27~M & 89.47 & 92.34 & 90.27 & 99.18 & 11.80 & $+2.48$ [$+2.35$, $+2.61$] & 50/50 \\
LoRA $r{=}16^{\ddagger}$ & \checkmark & -- & 4.33~M & 90.13$\pm$0.04 & 92.80$\pm$0.02 & 90.79$\pm$0.14 & 99.24$\pm$0.02 & 7.37$\pm$0.53 & $+1.82$ [$+1.71$, $+1.94$] & 50/50 \\
Conv-LoRA $r{=}16^{\ddagger}$ & \checkmark & -- & 4.35~M & 90.22$\pm$0.02 & 92.90$\pm$0.02 & 90.96$\pm$0.12 & 99.24$\pm$0.01 & 6.68$\pm$0.38 & $+1.73$ [$+1.63$, $+1.83$] & 50/50 \\
AdaptFormer & \checkmark & -- & 0.45~M & 89.52 & 92.53 & 90.68 & 99.15 & 11.18 & $+2.44$ [$+2.28$, $+2.59$] & 50/50 \\
Read-out only (ours) & -- & \checkmark & 0.20~M & 90.56 & 92.15 & \textbf{93.27} & 99.10 & 5.64 & $+1.40$ [$+1.18$, $+1.61$] & 50/50 \\
Conv-LoRA $r{=}16$ + read-out & \checkmark & \checkmark & 4.55~M & \underline{91.82} & \underline{93.89} & 92.50 & \underline{99.37} & \underline{3.02} & $+0.13$ [$+0.03$, $+0.23$]$^{\dagger}$ & 33/50 \\
LoRA $r{=}16$ + read-out & \checkmark & \checkmark & 4.54~M & 91.78 & \textbf{93.90} & \underline{92.74} & 99.34 & \textbf{3.00} & -- & -- \\
\textbf{RBA (ours)}$^{\ddagger}$ & \checkmark & \checkmark & 0.47~M & \textbf{91.93}$\pm$0.02 & 93.88$\pm$0.07 & 92.52$\pm$0.48 & \textbf{99.38}$\pm$0.04 & 3.03$\pm$0.44 & -- & -- \\
\bottomrule
\end{tabular}}
\end{table*}

Without an adapter, SEEM reached a Dice@4 of 75.83\%. Every encoder adapter raised it to about 90\%, whatever its size, and sixteen times the encoder rank added less than 0.7\%. Conv-LoRA gained only 0.09\% over LoRA of the same rank. RBA reached $91.93\pm0.02\%$ with 0.47~M parameters. On seed 42 it outperformed LoRA with rank 16, Conv-LoRA and AdaptFormer on all 50 test images, by 1.82\%, 1.73\% and 2.44\%. Over the three seeds its lead was 1.80\% over LoRA with rank 16 and 1.72\% over Conv-LoRA, with per-seed leads from 1.76\% to 1.82\% and from 1.71\% to 1.72\%. The margin was larger in IoU, at 85.16\% against 82.26\% for Conv-LoRA.

The read-out also improved every encoder adapter to which it was added, again on all 50 images, by 2.48\% on LoRA with rank 1, 1.65\% on LoRA with rank 16 and 1.60\% on Conv-LoRA. With the read-out in place, the size of the encoder adapter made little difference. RBA, which uses rank 1, and Conv-LoRA with the read-out differed by 0.13\% on 33 of 50 images. This difference was significant, with a corrected $p$-value of 0.012, but it lay inside the $\pm0.3\%$ margin, so the two were practically equivalent. The read-out branch alone outperformed Conv-LoRA by 0.33\% on 37 of 50 images ($p=0.008$).

\begin{figure}[]
\centering
\figfile{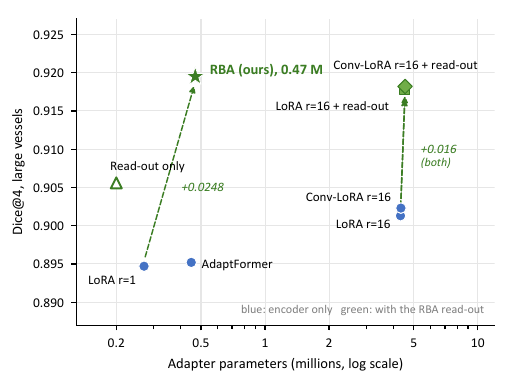}{0.95\linewidth}
\caption{Dice@4 on large vessels against adapter parameters on a logarithmic scale, for seed 42. Blue marks encoder-only adapters, green marks adapters with the RBA read-out, and dashed arrows mark its addition.}
\label{fig:tradeoff}
\end{figure}

RBA was not the best on every measure. LoRA with rank 16 and the read-out had a marginally higher clDice, 93.90\% against 93.88\%, and needed about as many clicks to reach 90\% Dice, 3.00 against 3.03, at nearly ten times the adapter size. The read-out branch alone had the highest sensitivity and the lowest specificity of the adapted rows, since it marked more pixels as vessel, both correctly and wrongly. Fig.~\ref{fig:tradeoff} plots Dice@4 against the number of adapter parameters. Every configuration with the read-out lay above every encoder-only adapter.

The decoder of SEEM, with 40.4~M parameters, was trained in every row of Table~\ref{tab:adapters}. Three rows were therefore repeated with all of SEEM frozen, so that only the adapter and the click prompt were trained, and Table~\ref{tab:frozen} compares them with the trained decoder. Freezing SEEM changed nothing measurable, since no change in Dice@4 exceeded 0.03\% or was significant. RBA then trained 1.04~M parameters, which are 0.30\% of SEEM, and still outperformed Conv-LoRA by 1.68\% on all 50 images, as it did with the decoder trained. With the click prompt also switched off, only the 0.47~M parameters of RBA were trained, which are 0.14\% of SEEM, and Dice@4 was 91.75\%. The difference from the row with the prompt was not significant.

\begin{table}[]
\centering
\caption{Effect of freezing the decoder of SEEM on large vessels (seed 42). Trained gives millions of trained parameters, and Change is frozen minus trained, in \%. In the last row the click prompt is also off.}
\label{tab:frozen}
\footnotesize
\setlength{\tabcolsep}{4pt}
\resizebox{\linewidth}{!}{%
\begin{tabular}{lccccc}
\toprule
 & \multicolumn{2}{c}{Decoder trained} & \multicolumn{2}{c}{SEEM frozen} & \\
\cmidrule(lr){2-3}\cmidrule(lr){4-5}
Adapter & Trained & Dice@4 & Trained & Dice@4 & Change \\
\midrule
LoRA $r{=}1$ & 40.65 & 89.47 & 0.84 & 89.44 & $-0.03$ \\
Conv-LoRA $r{=}16$ & 44.73 & 90.23 & 4.91 & 90.26 & $+0.03$ \\
RBA & 40.85 & \textbf{91.95} & 1.04 & \textbf{91.94} & $-0.01$ \\
RBA, click prompt off & -- & -- & 0.47 & 91.75 & -- \\
\bottomrule
\end{tabular}}
\end{table}

\subsection{Generalisation across targets and datasets}\label{sec:general}

The gain of RBA was then tested beyond the large vessels of the 3~mm subset. RBA was compared with Conv-LoRA, the strongest encoder adapter in Table~\ref{tab:adapters}, on every target of the OCTA-500 subsets and of ROSE-O. Table~\ref{tab:targets} lists Dice@4, clDice and NoC@90 of both models for each target, the paired difference in Dice@4 with its 95\% interval, the number of images on which RBA was better and the corrected $p$-value.

\begin{table*}[!t]
\centering
\caption{RBA against Conv-LoRA $r{=}16$ on every target, scores in \%. $\Delta$ is the paired change in Dice@4 with its 95\% interval for seed 42, and $p$ is Holm-corrected. $^{\ddagger}$Mean over three seeds. Bold marks the better Dice@4 where the difference is significant.}
\label{tab:targets}
\footnotesize
\setlength{\tabcolsep}{4pt}
\begin{tabular}{llccccccc}
\toprule
 & & \multicolumn{2}{c}{Dice@4} & & & & & \\
\cmidrule(lr){3-4}
Dataset & Target & RBA & Conv-LoRA & $\Delta$ [95\% CI] & Better & $p$ & clDice R / C & NoC@90 R / C \\
\midrule
\multirow{4}{*}{OCTA-500 3~mm} & Large vessels$^{\ddagger}$ & \textbf{91.93} & 90.22 & $+1.73$ [$+1.63$, $+1.83$] & 50/50 & $<10^{-10}$ & 93.88 / 92.90 & 3.03 / 6.68 \\
 & Veins & \textbf{90.49} & 88.87 & $+1.62$ [$+1.31$, $+1.92$] & 45/50 & $<10^{-10}$ & 92.78 / 92.28 & 6.72 / 13.54 \\
 & Arteries & \textbf{91.26} & 89.69 & $+1.57$ [$+1.30$, $+1.84$] & 48/50 & $<10^{-10}$ & 93.17 / 92.39 & 4.48 / 10.48 \\
 & FAZ & 97.52 & 97.86 & $-0.34$ [$-0.90$, $+0.21$] & 25/50 & 0.59 & -- & 1.00 / 1.00 \\
\midrule
\multirow{4}{*}{OCTA-500 6~mm} & Large vessels & \textbf{88.67} & 85.91 & $+2.76$ [$+2.59$, $+2.92$] & 50/50 & $<10^{-10}$ & 91.08 / 89.44 & 13.40 / 20.00 \\
 & Veins & \textbf{87.25} & 85.27 & $+1.99$ [$+1.70$, $+2.28$] & 47/50 & $<10^{-10}$ & 90.11 / 89.66 & 16.22 / 20.00 \\
 & Arteries & \textbf{87.56} & 84.57 & $+2.99$ [$+2.75$, $+3.24$] & 50/50 & $<10^{-10}$ & 90.01 / 88.37 & 17.04 / 20.00 \\
 & FAZ & 91.29 & 92.59 & $-1.30$ [$-2.78$, $+0.17$] & 23/50 & 0.084 & -- & 4.04 / 3.36 \\
\midrule
\multirow{2}{*}{ROSE-O} & Vessels & \textbf{77.55} & 76.23 & $+1.32$ [$+0.66$, $+1.99$] & 8/9 & 0.008 & 69.09 / 68.41 & 20.00 / 20.00 \\
 & FAZ & 95.24 & 95.90 & $-0.66$ [$-1.77$, $+0.45$] & 4/9 & 0.36 & -- & 1.00 / 1.00 \\
\bottomrule
\end{tabular}
\end{table*}

On the three vessel targets of the 3~mm subset the gain was stable, at 1.73\% on large vessels, 1.62\% on veins and 1.57\% on arteries, with corrected $p<10^{-10}$. The clDice was higher on all three, and the clicks needed to reach 90\% Dice fell by about half. On veins, RBA also outperformed Conv-LoRA with the same read-out, by 0.81\% on 42 of 50 images. On the FAZ, the difference of $-0.34\%$ was not significant, and its interval, from $-0.90\%$ to $+0.21\%$, was too wide to call the models equal.

The gain carried over to the other two datasets. On the 6~mm subset it was larger on every vessel target, at 2.76\% on large vessels, 1.99\% on veins and 2.99\% on arteries, with $p<10^{-10}$. A second training seed of RBA on 6~mm large vessels gave 88.98\%, against 88.67\% for seed 42. Conv-LoRA never reached 90\% Dice within 20 clicks on these targets, whereas RBA did on part of the images. The larger gain is consistent with the thinner vessels of the 6~mm subset, where a vessel spans about 1.5 times fewer pixels, since the gain of the read-out concentrated on thin vessels, as Section~\ref{sec:ablation} shows. On ROSE-O, RBA gained 1.32\% on vessels, on 8 of the 9 test subjects with $p=0.008$, although both models stayed well below their OCTA-500 accuracy. On the FAZ of the 6~mm subset and of ROSE-O, the differences of $-1.30\%$ and $-0.66\%$ were not significant.

The pixel measures used in the OCTA-500 paper complete this comparison. Table~\ref{tab:pixel} lists them, namely Dice, IoU, accuracy, sensitivity and specificity, for both models after four clicks. On every vessel target, RBA had the higher sensitivity and the higher specificity, by 0.5\% to 2.1\% and by 0.1\% to 0.7\%. On the FAZ the pattern reversed, as for a region drawn slightly too large. The IoU showed the same ordering as Dice with larger margins, for example 2.90\% on 3~mm large vessels. Accuracy differed by at most 0.6\%, because background pixels dominate it. Supplementary Tables S1 to S5 set these measures beside those of the OCTA-500 baselines, and Supplementary Tables S6 to S9 give them for every trained row.

\begin{table*}[!t]
\centering
\caption{Pixel measures of Conv-LoRA $r{=}16$ (C) and RBA (R) after four clicks, in \%. $^{\ddagger}$Mean over three seeds; the other rows use seed 42. Bold marks the better of the two.}
\label{tab:pixel}
\footnotesize
\setlength{\tabcolsep}{4.2pt}
\begin{tabular}{llcccccccccc}
\toprule
 & & \multicolumn{2}{c}{Dice} & \multicolumn{2}{c}{IoU} & \multicolumn{2}{c}{ACC} & \multicolumn{2}{c}{SE} & \multicolumn{2}{c}{SP} \\
\cmidrule(lr){3-4}\cmidrule(lr){5-6}\cmidrule(lr){7-8}\cmidrule(lr){9-10}\cmidrule(lr){11-12}
Dataset & Target & C & R & C & R & C & R & C & R & C & R \\
\midrule
\multirow{4}{*}{OCTA-500 3~mm} & Large vessels$^{\ddagger}$ & 90.22 & \textbf{91.93} & 82.26 & \textbf{85.16} & 98.69 & \textbf{98.92} & 90.96 & \textbf{92.52} & 99.24 & \textbf{99.38} \\
 & Veins & 88.87 & \textbf{90.49} & 80.03 & \textbf{82.69} & 99.35 & \textbf{99.44} & 90.36 & \textbf{91.24} & 99.62 & \textbf{99.69} \\
 & Arteries & 89.69 & \textbf{91.26} & 81.39 & \textbf{84.00} & 99.23 & \textbf{99.35} & 90.52 & \textbf{92.02} & 99.57 & \textbf{99.63} \\
 & FAZ & \textbf{97.86} & 97.52 & \textbf{95.88} & 95.32 & \textbf{99.84} & 99.80 & 98.58 & \textbf{98.89} & \textbf{99.89} & 99.83 \\
\midrule
\multirow{4}{*}{OCTA-500 6~mm} & Large vessels & 85.91 & \textbf{88.67} & 75.39 & \textbf{79.74} & 97.25 & \textbf{97.82} & 88.45 & \textbf{89.70} & 98.18 & \textbf{98.68} \\
 & Veins & 85.27 & \textbf{87.25} & 74.43 & \textbf{77.56} & 98.54 & \textbf{98.73} & 87.40 & \textbf{89.54} & 99.12 & \textbf{99.21} \\
 & Arteries & 84.57 & \textbf{87.56} & 73.36 & \textbf{77.98} & 98.63 & \textbf{98.92} & 87.51 & \textbf{89.02} & 99.13 & \textbf{99.37} \\
 & FAZ & \textbf{92.59} & 91.29 & \textbf{87.01} & 85.24 & \textbf{99.82} & 99.79 & 95.22 & \textbf{95.67} & \textbf{99.90} & 99.87 \\
\midrule
\multirow{2}{*}{ROSE-O} & Vessels & 76.23 & \textbf{77.55} & 61.86 & \textbf{63.61} & 90.96 & \textbf{91.58} & 76.38 & \textbf{76.85} & 94.58 & \textbf{95.27} \\
 & FAZ & \textbf{95.90} & 95.24 & \textbf{92.13} & 90.97 & \textbf{99.74} & 99.71 & 95.52 & \textbf{95.72} & \textbf{99.88} & 99.84 \\
\bottomrule
\end{tabular}
\end{table*}

\subsection{Comparison with published methods}\label{sec:published}

Published Dice scores were collected to place these results in context. They were not re-run, and the settings differ. The automatic networks use six projections, one projection or the OCTA volume. SAM-OCTA and SAM-OCTA2 receive prompt points drawn from the ground truth, whereas RBA receives four error-driven clicks and selects its own output. Table~\ref{tab:published} lists the published results on OCTA-500. The U-Net, Attention U-Net and IPN-V2 rows are the single-task results of the OCTA-500 paper~\cite{li2024octa}, and the IPN V2+, FARGO and Joint-Seg rows come from Table I of SAM-OCTA2~\cite{chen2025sam}. The other rows come from the original papers. Table~\ref{tab:rose} lists the published results on ROSE-O, which come from Table I of Hao et al.~\cite{hao2022retinal}, who used the three projections and 30 training subjects. D2Net is the exception, and its result comes from its own paper and uses the SVC projection only.

\begin{table*}[!t]
\centering
\caption{Published Dice scores on the OCTA-500 test sets in \%, for context. Results of other methods are taken from the cited papers. A dash marks values not reported, $^{\dagger}$ marks images 10200--10300 instead of 10251--10300, and $^{\ddagger}$ marks the mean over three seeds. Bold: best; underlined: second best.}
\label{tab:published}
\footnotesize
\setlength{\tabcolsep}{3.6pt}
\resizebox{\textwidth}{!}{%
\begin{tabular}{llcccccccc}
\toprule
 & & \multicolumn{2}{c}{Large vessels} & \multicolumn{2}{c}{Arteries} & \multicolumn{2}{c}{Veins} & \multicolumn{2}{c}{FAZ} \\
\cmidrule(lr){3-4}\cmidrule(lr){5-6}\cmidrule(lr){7-8}\cmidrule(lr){9-10}
Method & Setting & 3~mm & 6~mm & 3~mm & 6~mm & 3~mm & 6~mm & 3~mm & 6~mm \\
\midrule
U-Net~\cite{ronneberger2015u} & Automatic, six projections & 92.04 & 89.11 & 89.58 & 87.27 & 87.99 & 86.90 & 96.48 & 87.55 \\
Attention U-Net~\cite{oktay2018attention} & Automatic, six projections & 91.91 & 89.26 & 90.14 & \underline{87.35} & 88.07 & 87.03 & 97.76 & 86.73 \\
IPN-V2~\cite{li2024octa} & Automatic, OCTA volume & \underline{92.32} & 89.17 & \underline{90.78} & 87.27 & \underline{89.71} & \textbf{87.30} & 97.68 & 90.12 \\
IPN V2+~\cite{li2020ipn} & Automatic & \textbf{92.74} & \underline{89.41} & -- & -- & -- & -- & 97.55 & 90.84 \\
FARGO~\cite{peng2021fargo} & Automatic & 91.68 & 89.15 & -- & -- & -- & -- & 98.39 & \underline{92.72} \\
Joint-Seg~\cite{hu2022joint} & Automatic & 91.13 & \textbf{89.72} & -- & -- & -- & -- & \underline{98.43} & 90.51 \\
SSW-OCTA~\cite{chen2024snake} & Automatic & 92.15 & -- & -- & -- & -- & -- & 98.16 & -- \\
Melo et al.~\cite{melo2026multitask} & Automatic & -- & -- & -- & -- & -- & -- & \textbf{98.49} & -- \\
D2Net~\cite{hu2025dual} & Automatic, one projection & 91.23 & -- & -- & -- & -- & -- & -- & -- \\
LSENet~\cite{huang2026gaze} & Automatic, one projection & 92.03 & 89.14 & -- & -- & -- & -- & -- & -- \\
SAM-OCTA~\cite{chen2025samocta} & Label prompts, best setting & 91.57 & 88.98$^{\dagger}$ & 88.95 & 85.80$^{\dagger}$ & 88.72 & 85.99$^{\dagger}$ & 98.24 & 91.26$^{\dagger}$ \\
SAM-OCTA2~\cite{chen2025sam} & Label prompts & 92.07 & 89.23 & -- & -- & -- & -- & 98.33 & \textbf{92.84} \\
\midrule
\textbf{RBA (ours)} & 4 clicks, own selection & 91.93$^{\ddagger}$ & 88.67 & \textbf{91.26} & \textbf{87.56} & \textbf{90.49} & \underline{87.25} & 97.52 & 91.29 \\
\bottomrule
\end{tabular}}
\end{table*}

\begin{table}[!htb]
\centering
\caption{Published Dice scores on the ROSE-O test set in \%, for context. Bold: best; underlined: second best.}
\label{tab:rose}
\footnotesize
\setlength{\tabcolsep}{4pt}
\resizebox{\linewidth}{!}{%
\begin{tabular}{llcc}
\toprule
Method & Type & Vessels & FAZ \\
\midrule
U-Net~\cite{ronneberger2015u} & Single-task & 70.32 & 93.82 \\
OCTA-Net~\cite{ma2020rose} & Single-task & 75.83 & 94.64 \\
UTNet~\cite{gao2021utnet} & Single-task & 75.08 & 94.47 \\
FARGO~\cite{peng2021fargo} & Multi-task & 74.59 & 94.42 \\
MTI-Net~\cite{vandenhende2020mti} & Multi-task & 75.34 & 94.77 \\
VAFF-Net~\cite{hao2022retinal} & Multi-task & \underline{76.58} & \underline{95.19} \\
D2Net~\cite{hu2025dual} & Single-task, SVC & 75.37 & -- \\
\textbf{RBA (ours)} & Interactive, 4 clicks & \textbf{77.55} & \textbf{95.24} \\
\bottomrule
\end{tabular}}
\end{table}

On the 3~mm subset, RBA had the highest artery and vein Dice, at 91.26\% and 90.49\% against 90.78\% and 89.71\% for IPN-V2. On large vessels it was level with LSENet~\cite{huang2026gaze}, at 91.93\% against 92.03\%, and 0.81\% below IPN V2+, which reached 92.74\%. On the FAZ it stayed below the dedicated networks, at 97.52\% against up to 98.49\%. On the 6~mm subset, RBA had the highest artery Dice, 87.56\% against 87.35\%, and was level with the best vein result, 87.25\% against 87.30\%. Its large-vessel Dice of 88.67\% was below every published result, which range from 88.98\% to 89.72\%, and its FAZ Dice of 91.29\% was below the best, 92.84\%. On ROSE-O, RBA reached a vessel Dice of 77.55\%, above the best published value of 76.58\% for VAFF-Net, which used all 30 training subjects against 25 for RBA and no clicks. Its FAZ Dice was level with that of VAFF-Net, at 95.24\% against 95.19\%. The gain over Conv-LoRA on 6~mm large vessels is therefore a gain within SEEM and not over the published methods.

\subsection{Ablation and error analysis}\label{sec:ablation}

To separate the contributions of the two branches, RBA was trained with each branch alone and with both. Table~\ref{tab:ablation} lists the Dice@4 of the four combinations on large vessels and veins. On large vessels, either branch alone recovered most of the gap left by SEEM without an adapter, with 13.6\% for the encoder branch and 14.7\% for the read-out branch. On veins the read-out branch alone reached only 71.32\%, against 87.98\% for the encoder branch alone. On top of the encoder branch, the read-out added 2.5\% on both targets, on every test image. This gain was smaller than the gain of the read-out without the encoder branch, by 12.2\% on large vessels and 17.8\% on veins, with 95\% intervals of $[-12.8, -11.7]$ and $[-18.9, -16.8]$. The effects of the two branches therefore overlapped and were not complementary.

\begin{table}[!htb]
\centering
\caption{Ablation of the two branches of RBA, Dice@4 in \% on 50 test images (seed 42), with paired changes and the interaction. The decoder of SEEM is trained in every row.}
\label{tab:ablation}
\footnotesize
\setlength{\tabcolsep}{4pt}
\resizebox{\linewidth}{!}{%
\begin{tabular}{lcccc}
\toprule
Encoder branch & Read-out branch & Params (M) & Large vessels & Veins \\
\midrule
-- & -- & 0 & 75.83 & 50.98 \\
\checkmark & -- & 0.27 & 89.47 & 87.98 \\
-- & \checkmark & 0.20 & 90.56 & 71.32 \\
\checkmark & \checkmark & 0.47 & \textbf{91.95} & \textbf{90.49} \\
\midrule
\multicolumn{3}{l}{Read-out added to encoder} & $+2.48$ & $+2.51$ \\
\multicolumn{3}{l}{Encoder added to read-out} & $+1.40$ & $+19.16$ \\
\multicolumn{3}{l}{Interaction} & $-12.2$ & $-17.8$ \\
\bottomrule
\end{tabular}}
\end{table}

The weakness of the read-out branch on veins concerned vessel identity and not detection. Table~\ref{tab:identity} lists the share of vein pixels that were found, the share of artery pixels marked as vein and the share of false positives on arteries. The read-out branch alone found 85.7\% of the vein pixels, close to the other rows, but it marked 30.4\% of the artery pixels as vein, and about 70\% of its false positives lay on arteries. With any encoder branch, about 2\% of the artery pixels were marked as vein.

\begin{table}[!htb]
\centering
\caption{Vessel identity on the vein target, in \% (seed 42). Found is the share of vein pixels found, Art.\ as vein the share of artery pixels marked as vein, and FP on art.\ the share of false positives on arteries.}
\label{tab:identity}
\footnotesize
\setlength{\tabcolsep}{4pt}
\resizebox{\linewidth}{!}{%
\begin{tabular}{lcccc}
\toprule
Row & Dice@4 & Found & Art.\ as vein & FP on art. \\
\midrule
No adapter & 50.98 & 64.5 & 41.3 & 59.8 \\
Read-out only & 71.32 & 85.7 & 30.4 & 69.5 \\
LoRA $r{=}1$ & 87.98 & 89.6 & 2.1 & 9.4 \\
Conv-LoRA $r{=}16$ & 88.87 & 90.3 & 1.8 & 7.4 \\
Conv-LoRA + read-out & 89.68 & 93.0 & 2.4 & 12.1 \\
RBA (ours) & 90.49 & 91.2 & 2.3 & 15.1 \\
\bottomrule
\end{tabular}}
\end{table}

The gain of the read-out concentrated on thin vessels and edges. Table~\ref{tab:calibre} splits the large-vessel results by vessel width and by distance to the vessel edge. Thin vessels, below three pixels wide, make up 24\% of the vessel pixels, and thick vessels are seven pixels wide or more. Adding the read-out to LoRA with rank 1 raised thin-vessel recall from 77.6\% to 84.0\% and centre-line recall from 90.3\% to 93.5\%. It also lowered the error within one pixel of the vessel edge from 11.7\% to 8.4\%, each on all 50 test images. Thick-vessel recall changed by only 0.2\%, and on Conv-LoRA the pattern was the same. One trade-off was visible. A larger share of the remaining false positives lay more than 2 pixels from any vessel, at 19\% for RBA against 9\% to 11\% for the encoder-only adapters.

\begin{table*}[!t]
\centering
\caption{Large-vessel results by vessel width and edge proximity, in \% (seed 42). Lower is better for the edge error and for distant false positives (FP). The last rows give the paired change when the read-out is added, with the number of images that improved. Bold marks the best value.}
\label{tab:calibre}
\footnotesize
\resizebox{\textwidth}{!}{%
\begin{tabular}{lccccccc}
\toprule
Row & Dice@4 & Recall, thin & Recall, mid & Recall, thick & Edge error & Centre-line recall & Distant FP \\
\midrule
No adapter & 75.83 & 52.5 & 88.7 & 94.9 & 24.8 & 76.8 & 18.5 \\
LoRA $r{=}1$ & 89.47 & 77.6 & 94.3 & 97.2 & 11.7 & 90.3 & \textbf{9.4} \\
Conv-LoRA $r{=}16$ & 90.23 & 79.5 & 94.5 & 97.4 & 10.8 & 91.2 & 11.0 \\
Read-out only & 90.56 & \textbf{85.8} & \textbf{95.6} & 97.5 & 9.0 & \textbf{94.1} & 29.3 \\
Conv-LoRA + read-out & 91.82 & 84.2 & 95.1 & \textbf{97.7} & 8.6 & 93.5 & 17.8 \\
RBA (ours) & \textbf{91.95} & 84.0 & 94.9 & 97.4 & \textbf{8.4} & 93.5 & 19.2 \\
\midrule
Read-out added to LoRA $r{=}1$ & & $+6.3$ (50/50) & & $+0.2$ & $-3.4$ (50/50) & $+3.2$ (50/50) & \\
Read-out added to Conv-LoRA & & $+4.6$ (48/50) & & $+0.3$ & $-2.2$ (50/50) & $+2.3$ (49/50) & \\
\bottomrule
\end{tabular}}
\end{table*}

A second, much smaller read-out design, called RiDA here, was also tested. It uses rank-1 codes sharpened by a fixed Hessian-based ridge operator~\cite{frangi1998multiscale} and has 0.29~M parameters. It did not outperform RBA, as Supplementary Table S9 reports, but its read-out alone still added 1.78\% to LoRA with rank 1.

These results are consistent with the structure of the read-out in Eq.~\eqref{eq:readout}. In SEEM, every candidate mask is a weighted sum of the same 512 maps at a quarter of the input resolution. An encoder adapter changes the content of these maps but not their resolution, whereas RBA adds high-resolution directions to them. Accordingly, the gain was stable on thin, elongated targets and absent on the FAZ, a large smooth region for which a quarter-resolution basis seems to suffice. Within the vessel targets, the gain landed on thin vessels and edges, while thick vessels were left almost unchanged, and RiDA showed the same effect.

The read-out branch alone failed on veins and arteries, where the model must decide which vessel type a pixel belongs to. A possible reason is that this decision needs context, such as how a vessel branches and where it crosses others, which only adapted encoder features carry. A rank-1 encoder branch was enough for this role once the read-out was adapted. LoRA with rank 16 and Conv-LoRA with the same read-out were not better on large vessels, and Conv-LoRA was even worse on veins, at 89.68\% against 90.49\%. In this work, it was therefore found that adaptation capacity in SEEM is better spent where the features are read than where they are computed.

\subsection{Qualitative results}\label{sec:qualitative}

\begin{figure}[]
\centering
\figfile{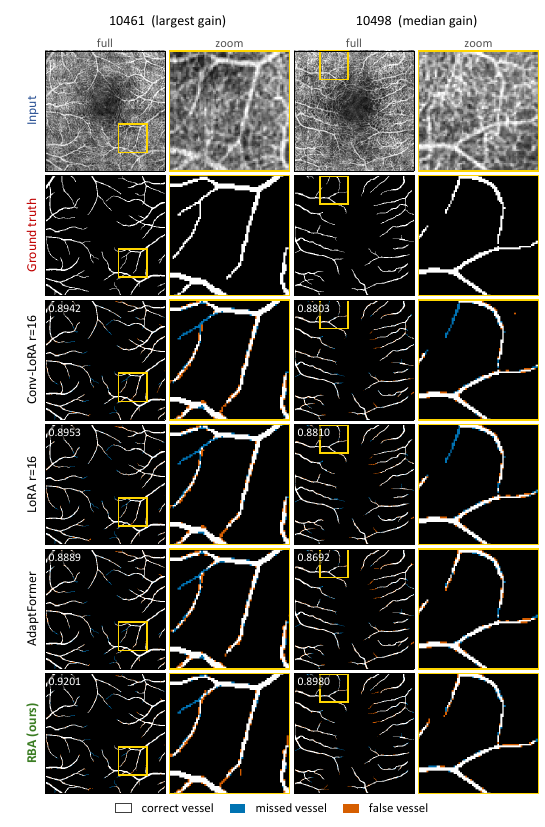}{0.95\linewidth}
\caption{Large-vessel segmentation after four clicks on test image 10461, with the largest gain, and image 10498, with the median gain. White marks correct vessel, blue missed vessel and orange false vessel. Numbers are Dice scores.}
\label{fig:qual_lv}
\end{figure}

Qualitative examples show where the gain of RBA appears in the images. For large vessels, Fig.~\ref{fig:qual_lv} shows the test images with the largest and the median gain of RBA over Conv-LoRA, images 10461 and 10498. Each image is shown in full and with a zoom on the area where the models disagreed most. On image 10461, Dice rose from 89.42\% with Conv-LoRA to 92.01\% with RBA, and on image 10498 from 88.03\% to 89.80\%, close to the mean gain in Table~\ref{tab:adapters}. In both zooms the gain came from thin branches. Conv-LoRA and LoRA with rank 16 missed a thin vessel on image 10498 that RBA and AdaptFormer recovered. On image 10461, all three encoder-only adapters missed most of a thin branch that RBA recovered in part. The remaining errors of all models were mostly one-pixel deviations along the edges of larger vessels, consistent with Table~\ref{tab:calibre}. The images were selected by their gain and illustrate the tables, but they do not by themselves show general performance.

\begin{figure*}[!t]
\centering
\figfile{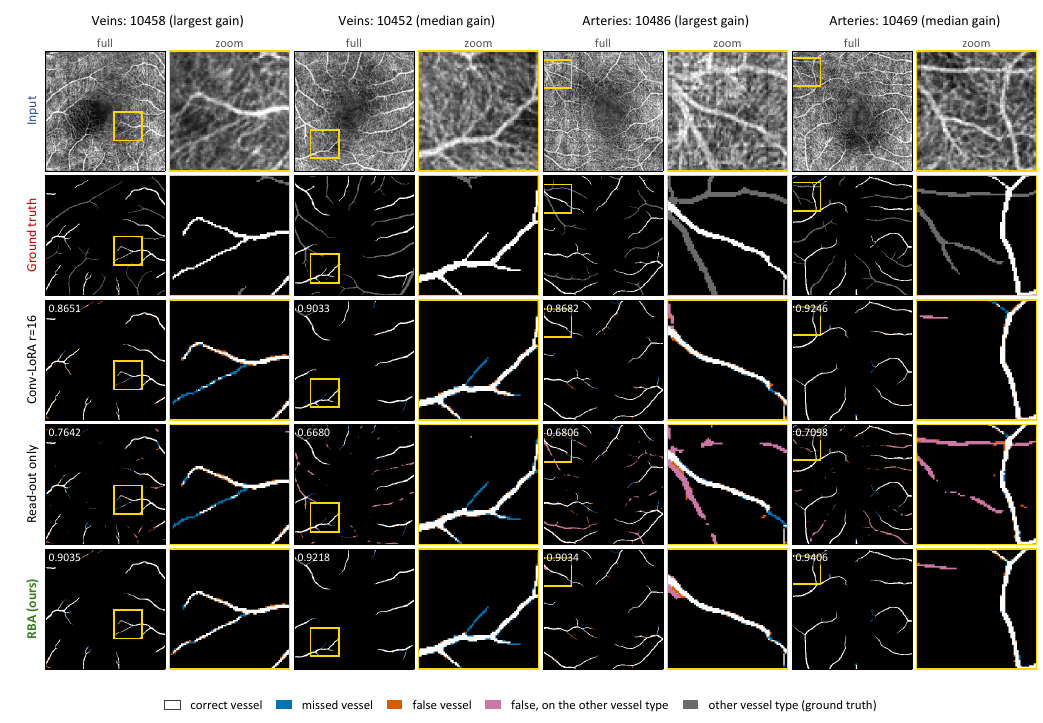}{\textwidth}
\caption{Vein and artery segmentation after four clicks on the images with the largest and the median gain. Colours as in Fig.~\ref{fig:qual_lv}, with pink for false vessel on the other vessel type and grey for the other vessel type in the ground truth. Numbers are Dice scores.}
\label{fig:qual_av}
\end{figure*}

The same comparison for veins and arteries is shown in Fig.~\ref{fig:qual_av}, with the read-out branch alone added as a further row. On the vein image with the largest gain, image 10458, Conv-LoRA missed most of a thin branch that RBA recovered, and Dice rose from 86.51\% to 90.35\%. On the artery image with the largest gain, image 10486, Dice rose from 86.82\% to 90.34\%, and the two median-gain images, 10452 and 10469, gained 1.85\% and 1.60\%. Without the encoder branch, the model marked many vessels of the other type as the target, shown in pink, and its Dice fell to between 67\% and 76\%, in line with Table~\ref{tab:identity}.

\subsection{Interaction behaviour and computational cost}\label{sec:interaction}

The gains in NoC@90 came mainly from better masks after the first click and not from larger gains per click. On large vessels, Dice@4 exceeded Dice@1 by only 0.04\% to 0.06\% for most adapted rows, and by 0.01\% to 0.05\% in the extra seeds of LoRA with rank 16 and Conv-LoRA. A click placed at random inside the error region lowered the Dice of RBA by only 0.04\% at the fourth click, so the gains did not rest on ideal click placement. On the FAZ, the Dice of RBA decreased slightly from 97.92\% at the first click to 97.52\% at the fourth, which was not significant.

The cost of the adapters was measured at test time on the 3~mm subset, as the mean over 48 test images after two warm-up images and without data loading. The encoder time covers the pass run once per image, including the RBA basis, and the image time adds four click rounds. Table~\ref{tab:cost} also lists the trained parameters of each row and the peak GPU memory. With four clicks, every row took 381 to 404~ms per image and 72 to 75~ms per click round, a time dominated by the decoder of SEEM. RBA took 401~ms against 400~ms for Conv-LoRA, and it needed 1.78~GB of GPU memory against 1.71~GB. The basis alone added about 7~ms to the encoder pass, which took 99~ms against 92~ms without an adapter. These properties make RBA practical for interactive use. The basis is computed once per image, so a click costs almost the same as with Conv-LoRA, and every model in this study was trained and tested on one laptop GPU with 8~GB.

\begin{table}[!htb]
\centering
\caption{Cost at test time on the 3~mm subset, large vessels, on one NVIDIA GeForce RTX 3070 Ti Laptop GPU. Encoder is the pass run once per image, and Image adds four click rounds. Trained is in millions of parameters, and Memory is peak GPU memory.}
\label{tab:cost}
\footnotesize
\setlength{\tabcolsep}{3pt}
\resizebox{\linewidth}{!}{%
\begin{tabular}{lccccc}
\toprule
Method & \makecell{Trained\\(M)} & \makecell{Memory\\(GB)} & \makecell{Encoder\\(ms)} & \makecell{Click\\(ms)} & \makecell{Image\\(ms)} \\
\midrule
No adapter & 40.38 & 1.70 & 92 & 74.6 & 390 \\
LoRA $r{=}1$ & 40.65 & 1.70 & 103 & 72.3 & 393 \\
LoRA $r{=}16$ & 44.71 & 1.71 & 105 & 72.3 & 394 \\
Conv-LoRA $r{=}16$ & 44.73 & 1.71 & 111 & 72.2 & 400 \\
AdaptFormer & 40.83 & 1.70 & 94 & 71.6 & 381 \\
Read-out only & 40.58 & 1.78 & 99 & 72.3 & 389 \\
LoRA $r{=}16$ + read-out & 44.92 & 1.80 & 111 & 72.8 & 402 \\
Conv-LoRA + read-out & 44.93 & 1.80 & 115 & 72.2 & 404 \\
RBA (ours) & 40.85 & 1.78 & 109 & 72.8 & 401 \\
RBA, SEEM frozen & 1.04 & 1.78 & 109 & 73.2 & 402 \\
\bottomrule
\end{tabular}}
\end{table}

\section{Conclusion}\label{sec:conclusion}

In this work, the Read-out Basis Adapter (RBA) was introduced to adapt the basis from which SEEM generates its candidate masks, rather than only its image encoder, for interactive retinal vessel segmentation in OCTA. With 0.47~M additional parameters, RBA starts exactly as SEEM and keeps its candidate generation and selection. It outperformed LoRA, Conv-LoRA and AdaptFormer on all 50 large-vessel test images of the OCTA-500 3~mm subset and, averaged over three training seeds, improved Dice by 1.72 percentage points over Conv-LoRA. The gain was also observed for veins and arteries, on the 6~mm subset and on ROSE-O, and it came mainly from thin vessels and vessel boundaries. The rank-1 encoder branch supplied vessel identity, which the read-out branch alone could not provide for arteries and veins. With the original SEEM components frozen, RBA kept its accuracy while training only 0.30\% of the parameters of SEEM. No significant difference was observed in FAZ segmentation, which suggests that the benefit of RBA lies in thin tubular structures. These findings show that, in SEEM, adaptation capacity pays more at the mask read-out than in the encoder.

Several directions can extend these findings. Multi-seed evaluations on the remaining targets and datasets, together with images from other OCTA devices, would establish the robustness and cross-device generalisability of the approach. Clicks placed along vessel boundaries or outside the error regions, followed by a study with expert annotators, would show how far the gains carry over to real interaction. Ablations of the number, sources and resolution of the basis maps would clarify which components contribute most to the gain. Extending the read-out adaptation to compact structures such as the FAZ is a further direction. Beyond SEEM, SAM and SAM~2 also form their masks from output tokens and one image embedding, so read-out adaptation can be investigated in these models. Finally, click-conditioned features could make each click more informative and strengthen the refinement over successive clicks.

\sloppy
\section*{CRediT authorship contribution statement}
\textbf{Jaafar M. Al-Ghabban:} Conceptualization, Methodology, Software, Validation, Formal analysis, Investigation, Data curation, Visualization, Writing -- original draft. \textbf{Mohammad Reza Mohammadi:} Conceptualization, Methodology, Supervision, Project administration, Writing -- review \& editing.

\section*{Declaration of competing interest}
The authors declare that they have no known competing financial interests or personal relationships that could have appeared to influence the work reported in this paper.

\section*{Declaration of generative AI and AI-assisted technologies in the manuscript preparation process}
The authors did not use generative AI or AI-assisted technologies to produce the scientific content, experiments, or results reported in this work. Where such tools were used to assist with language editing and drafting code help, the authors reviewed and edited the output and take full responsibility for the content of the published article.

\section*{Funding}
This research did not receive any specific grant from funding agencies in the public, commercial, or not-for-profit sectors.

\section*{Data availability}
OCTA-500~\cite{li2024octa} is publicly available at \url{https://ieee-dataport.org/open-access/octa-500}. ROSE and ROSE-O~\cite{ma2020rose,hao2022retinal} are available from their authors on request. Code for RBA and all analyses: \url{https://github.com/jaafaralghabban/SEEM-adapter}.

\section*{Acknowledgements}
The authors thank the School of Computer Engineering, Iran University of Science and Technology, for computational support, and the providers of OCTA-500 and ROSE-O for sharing their data.

\bibliographystyle{model1-num-names}
\bibliography{cas-refs}

\end{document}